\documentclass{scrartcl}

\usepackage[utf8]{inputenc}
\usepackage[english]{babel}
\usepackage{amsfonts, amsmath, amssymb}
\usepackage{graphicx}
\usepackage{xcolor}
\usepackage[colorlinks = True, linkcolor = red, citecolor = blue]{hyperref}
\usepackage{natbib}
\usepackage{url}
\usepackage{booktabs}
\usepackage{nicefrac}
\usepackage{microtype}
\usepackage[linesnumbered, ruled, boxed]{algorithm2e}
\usepackage{tikz}
\usepackage{array}
\usepackage{indentfirst}
\usepackage[title]{appendix}

\definecolor{myc1}{rgb}{1, 0, 0}
\definecolor{myc2}{rgb}{0.75, 0, 0.75}

\DeclareMathOperator*{\argmin}{arg\,min}
\DeclareMathOperator*{\argmax}{arg\,max}
\newcommand{\x}{\mathbf{x}}
\newcommand{\tx}{\tilde{\mathbf{x}}}
\newcommand{\ty}{\tilde{y}}
\newcommand{\dd}{\mathrm{d}}
\newcommand{\Hh}{\mathcal{H}}
\newcommand{\Db}{D_{\text{best}}}

\title{Targeted Active Learning for Bayesian Decision-Making}

\author{
Louis Filstroff\,$^1$ \hspace{0.5em} Iiris Sundin\,$^1$ \hspace{0.5em} Petrus Mikkola\,$^1$ \\
Aleksei Tiulpin\,$^1$ \hspace{0.5em} Juuso Kylmäoja\,$^1$ \hspace{0.5em} Samuel Kaski\,$^{1,2}$ \\
\small $^1$ Department of Computer Science, Aalto University, Finland \\
\small $^2$ Department of Computer Science, University of Manchester, UK
}

\begin{document}

\maketitle

\begin{abstract}
Active learning is usually applied to acquire labels of informative data points in supervised learning, to maximize accuracy in a sample-efficient way. However, maximizing the accuracy is not the end goal when the results are used for decision-making, for example in personalized medicine or economics. We argue that when acquiring samples sequentially, separating learning and decision-making is sub-optimal, and we introduce an active learning strategy which takes the down-the-line decision problem into account. Specifically, we introduce a novel active learning criterion which maximizes the expected information gain on the posterior distribution of the optimal decision. We compare our targeted active learning strategy to existing alternatives on both simulated and real data, and show improved performance in decision-making accuracy.
\end{abstract}

\section{Introduction} \label{sec-intro}

Supervised learning techniques aim at learning a function that maps the input $\x \in \mathcal{X}$ to the outcome (or label) $y \in \mathcal{Y}$, based on a collection of examples $\mathcal{D} = \{(\x_i, y_i)\}_{i=1}^{N}$. Whereas having access to thousands of unlabeled data is nowadays easy, obtaining the associated labels is expensive in many applications, such as those involving human experts (e.g., image annotating), or running additional experiments. In this context, \textit{active learning} (AL) aims at iteratively querying for the most informative data point among a pool of unlabeled data \citep{settles2012active}. In the machine learning literature, the term ``active learning'' often implies a classification task, but the concept straightforwardly extends to regression, and the same problem arises in the statistics literature under the name ``optimal experimental design'' or ``Bayesian experimental design'' \citep{chaloner1995bayesian,ryan2016review}.

Active learning then boils down to the selection criterion for the next point to label. Popular strategies include uncertainty sampling \citep{lewis1994heterogeneous}, expected error reduction \citep{roy2001toward}, or expected information gain (\citet{mackay1992information}; see detailed discussion in Section~\ref{sec-bal}). All these strategies aim at learning a model as accurate as possible with as few queries as possible. However, the accuracy of the model is not the end goal in all scenarios. In this paper, we consider the setup where the model is subsequently used for decision-making, i.e., where a user has to choose an action among a set of $K$ available ones. Such a problem arises in personalized medicine: based on the history of previous patients, described by patient covariates, the treatment they received, and the observed outcome, a doctor will use the model predictions to choose the best treatment for a new patient \citep{shalit2017estimating,alaa2017bayesian,bica2020from}. In such a setting, decisions are assessed by their so-called utility (e.g., balancing the effectiveness and side-effects of each treatment), and the optimal decision is the one which yields the highest expected utility.

Traditionally, model learning and decision-making are carried out separately, i.e., the learning phase is blind to the decision-making problem. This is not optimal when data can be collected actively, and there is a need for active learning strategies which take into account this hierarchical structure. This problem of \textit{decision-making-aware} active learning has recently received attention by \citet{sundin2019active}, who proposed a heuristic strategy for a binary decision-making problem. However, their criterion does not extend to more complex situations, such as multiple-decision problems, which limits applications.

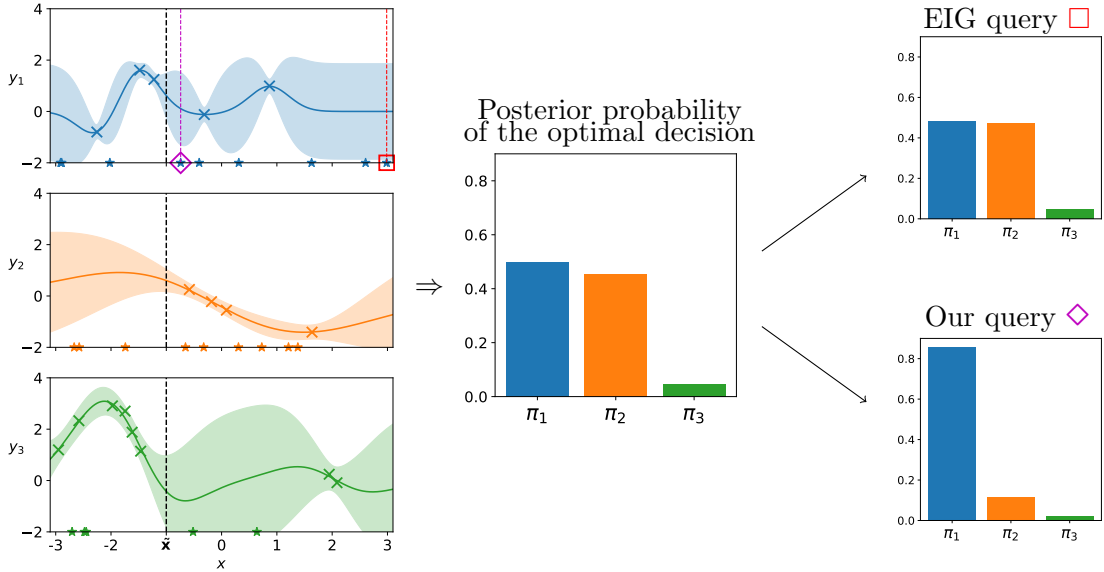
\begin{figure*}
    \centering
    \begin{tikzpicture}
        \node[inner sep=0pt] at (0,0)
            {\includegraphics[height=8cm]{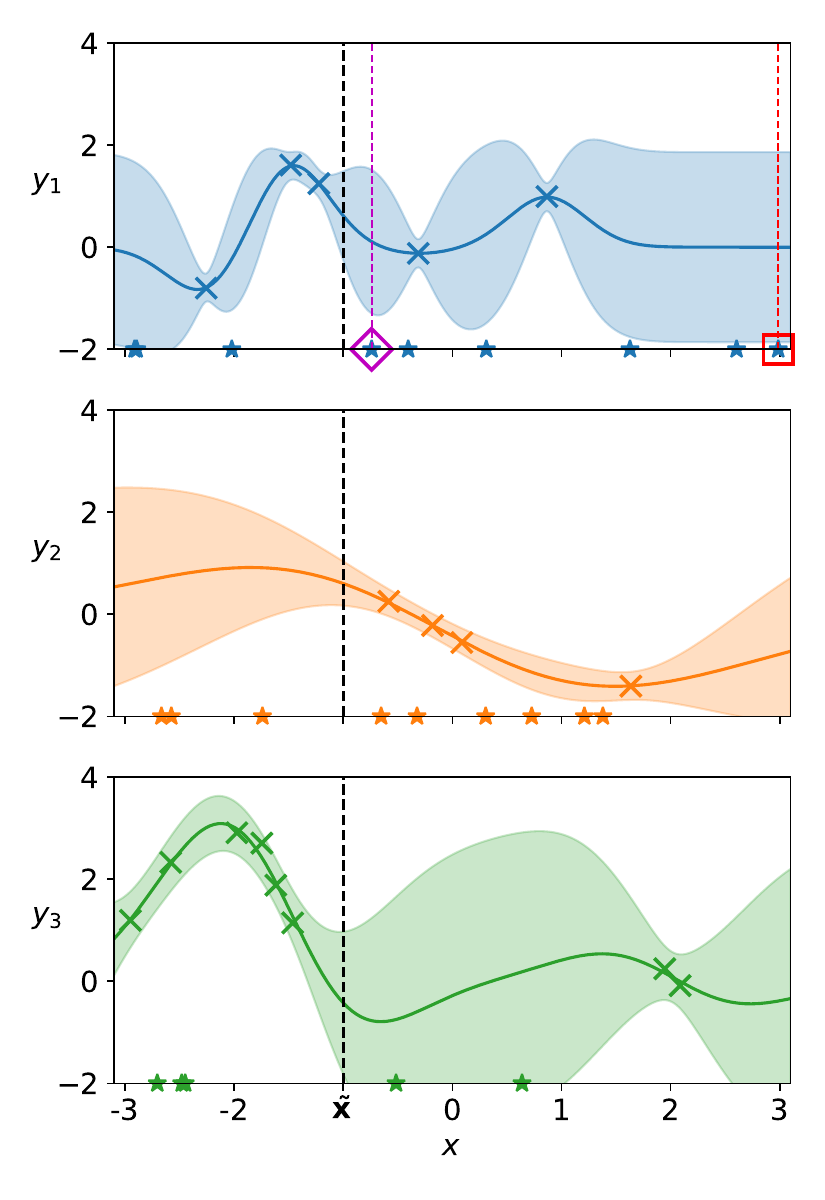}};
        
        \draw (3,0) node {$\Rightarrow$};
        
        \node[inner sep=0pt] at (5.25,0)
            {\includegraphics[height=4cm]{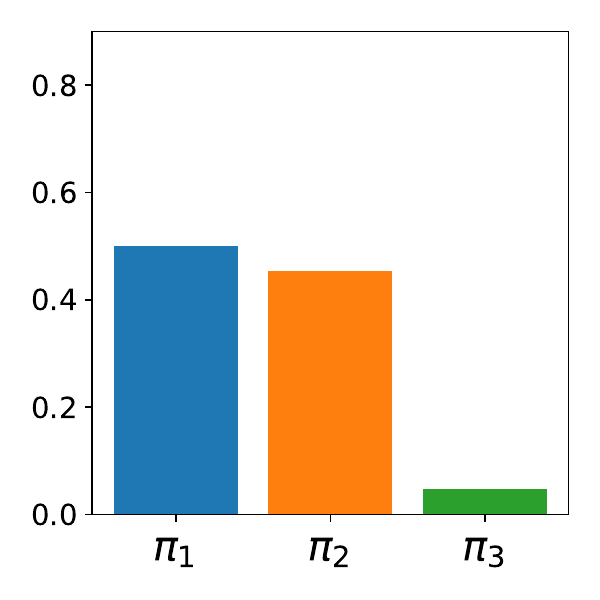}};
            
        \draw (5.375,2.35) node {Posterior probability};
        \draw (5.375,2.05) node {of the optimal decision};
        
        \draw [->] (7.375,0.5) -- (8.75,1.5);
        \draw [->] (7.375,-0.5) -- (8.75,-1.5);
        
        \node[inner sep=0pt] at (10.5,2)
            {\includegraphics[height=3cm]{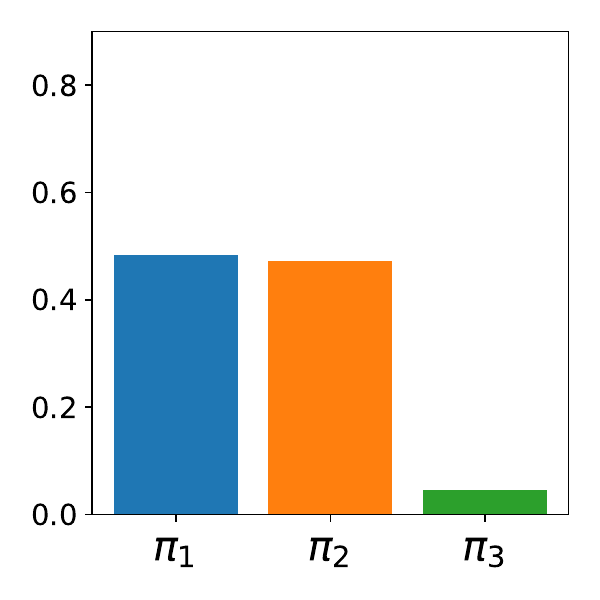}};
        
        \node[inner sep=0pt] at (10.5,-2)
            {\includegraphics[height=3cm]{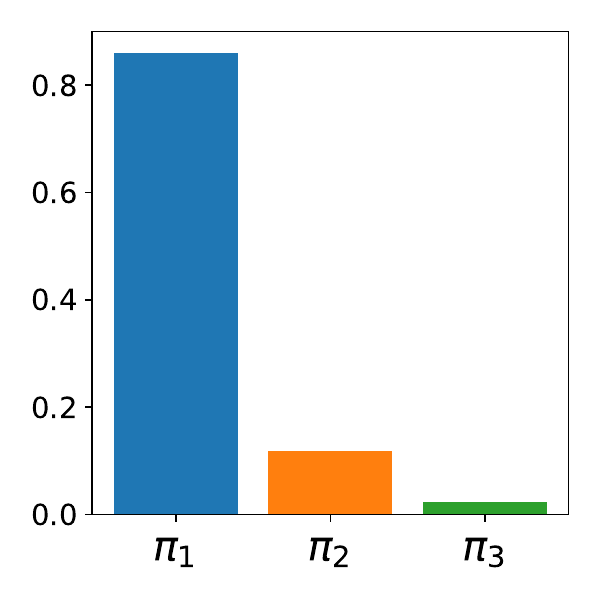}};
            
        \draw (10.6, 3.55) node {EIG query \textcolor{myc1}{$\square$}};
        \draw (10.6, -0.4) node {Our query \textcolor{myc2}{\LARGE $\diamond$}};
    \end{tikzpicture}
    \caption{Illustrative example of a decision-making task: choose one decision from $d=\{1,2,3\}$ at point $\tx = -1$ (black dashed line). (Left): Based on a learning dataset, outcome prediction functions $f_k$ have been computed (colored lines with uncertainty intervals). The learning dataset consists of labeled ($\times$ marker) and unlabeled ($\star$ marker) data points. (Center): The posterior distribution of the optimal decision helps in making the decision (the Bayes-optimal decision is $d=1$), and assessing its uncertainty. (Right): Evolution of that distribution after querying one additional point. Using the standard EIG criterion (Sec.~\ref{sec-bal}) does not help the decision-making (top), while the proposed targeted AL criterion greatly improves it by reducing its uncertainty (bottom).}
    \label{fig-1}
\end{figure*}

In this paper, we propose a principled selection criterion for decision-making-aware active learning. This criterion extends the classical expected information gain of Bayesian experimental design; more precisely, instead of applying it to the posterior distribution of the parameters of the model, as is usually done, we apply it to the \emph{posterior distribution of the optimal decision}, the distribution of interest in our case, as illustrated in Figure~\ref{fig-1}. We specifically consider \emph{performance improvement at first $t$} additional acquisitions for small $t$, which is a key measure for scenarios where acquisitions are costly, such as personalized medicine and generally having a human in the loop. Specifically, our method is applicable when a) the test population of the decision-making task is known: a single individual $\tx$ (as in personalized medicine), or a set of individuals, and b) it is possible to collect more data on-demand, for example once the model has been deployed, but c) these queries are costly due to, e.g., requesting new experiments, involving experts, or fulfilling privacy constraints. We empirically demonstrate the advantages of the proposed method with respect to existing AL baselines, both on simulated and real-world experiments.

\section{Problem formulation} \label{sec-problem}

\subsection{Modeling of outcomes}

We consider a regression setting with covariates $\x \in \mathbb{R}^{p}$ and outcomes $y \in \mathbb{R}$. We further assume that the outcome also depends on a decision variable $d \in \{1, \dotsc K \}$. Typically, the outcome is observed after an action has been taken. In the healthcare application, this corresponds to observing the effect of treatment $d$ on a patient. We therefore have a training set $\mathcal{D}$ comprising triplets, i.e., $\mathcal{D} = \{ (x_i, d_i, y_i) \}_{i=1}^{N}$.

Denoting by $y_k$ the variable $y|d = k$, the goal is therefore to learn the functions $f_k$ which map $\x$ to $y_k$. In this work, we assume that the $y_k$ are conditionally independent given $\x$, and write
\begin{equation}
    y_k = f_k(\x) + \epsilon_k,
\end{equation}
where $\epsilon_k \sim \mathcal{N}(0,\sigma_k^2)$.

Moreover, we assume that we are equipped with a functional prior distribution on $f_k$, such as a Gaussian process (GP) or a Bayesian neural network, which allows us to deal with posterior uncertainty. Indeed, given $\mathcal{D}_k = \{ (x_i, d_i, y_i) \in \mathcal{D}~|~d_i = k \}$, and using the notation $f_{k,\x}$ to denote $f_k(\x)$, we may characterize the posterior distribution $p(f_{k,\x}|\mathcal{D}_k)$ for all $\x$.

\subsection{Decision-making problem}

For clarity of presentation, the paper will focus on a single test input, which we denote by $\tx$, rather than a test population. Nevertheless, all developments presented in the paper straightforwardly extend to a test population; this is summarized in Eq.~\eqref{eq-crit-mult} and further details can be found in Appendix~\ref{app-E}.

The input $\tx$ is a previously unseen data point for which the end-user of the model has to make a decision, i.e., has to choose one among the set of the $K$ available decisions. In our introductory example, $\tx$ corresponds to the covariates of a patient for whom the doctor has to choose a treatment.

In this context, decisions are assessed through a scalar utility: the higher the better. Utilities can be computed from the outcomes $\ty_k$; we write $u_k = r_k(\ty_k)$, where the $r_k$ are known, deterministic functions that map the outcomes to the utilities. In the remainder of the paper we will assume, without loss of generality, that $u = y$.

Given that the models have been trained on $\mathcal{D} = \cup_k \mathcal{D}_k$, we assume that the user behaves optimally in the sense of (evidential) decision theory, i.e., chooses the decision which yields the greatest expected utility at $\tx$. The Bayes-optimal decision is
\begin{equation}
    d_{\text{BAYES}} = \argmax_{k \in \{ 1, \dotsc, K \}} \iint \ty_k p(\ty_k|f_{k,\tx}) p(f_{k,\tx}|\mathcal{D}_k) \mathrm{d} f_{k,\tx} \mathrm{d} \ty_k.
    \label{eq-dbayes}
\end{equation}

\subsection{Active learning} \label{sec-bal}

We assume access to a pool of unlabeled data ${\mathcal{U} = \{(\x_j, d_j)\}_{j=1}^{J}}$, from which the associated outcomes can be actively queried. We wish to select queries from $\mathcal{U}$ which are maximally useful for the decision-making problem, i.e., queries which  reduce uncertainty on the optimal decision for $\tx$.

\paragraph{Conventional Bayesian active learning.}

In a standard Bayesian regression formulation, i.e. when the relationship between the input $\x$ and the outcome $y$ is modeled by a likelihood $p(y|\x,\theta)$, where $\theta$ are latent parameters with a prior distribution $p(\theta)$, the active learning problem is referred to as Bayesian experimental design. As first suggested by \citet{lindley1956measure}, the optimal data point to query from an information-theoretic perspective (which we denote by $\x^{\star}$) is the one which maximizes the so-called expected information gain (EIG) on the posterior distribution of the parameters,
\begin{align}
    \x^{\star} = \argmax_{\x_j \in \mathcal{X}} & \bigg( \Hh[p(\theta|\mathcal{D})] \label{eq-eig} - \mathbb{E}_{p(y_j|\x_j, \mathcal{D})} \big[ \Hh[p(\theta|\mathcal{D} \cup \{ (\x_j, y_j) \}) ] \big] \bigg).
\end{align}
Here the notation $\Hh[p(.)]$ denotes the differential entropy of a probability distribution \citep{shannon1948mathematical}. This idea has then been considered by several authors, see for example \citet{bernardo1979expected, mackay1992information,houlsby2011bayesian,hernandez2014predictive}.

Note that the criterion of Eq.~\eqref{eq-eig} can be rearranged in a form which computes entropies in the outcome space rather than the parameter space (this has been coined ``BALD'' by \citet{houlsby2011bayesian}), which allows to define it in a non-parametric setting:
\begin{equation}
    \x^{\star} = \argmax_{\x_j \in \mathcal{X}} \bigg( \Hh[p(y|\x,\mathcal{D})] - \mathbb{E}_{p(f|\mathcal{D})} \big[ \Hh[p(y|\x, f) ] \big] \bigg),
    \label{eq-bald}
\end{equation}
see details on the equivalence in Appendix~\ref{app-C}. Nonetheless, the EIG remains a challenging criterion to compute, as it involves so-called nested Monte Carlo estimation \citep{rainforth2018nesting}, and several recent works have aimed at mitigating this issue \citep{foster2019variational,zheng2020sequential}.

\paragraph{Shortcomings for the decision-making problem.}

The conventional Bayesian AL criterion of Eq.~\eqref{eq-bald} could easily be adapted to our setting, and would select the element of $\mathcal{U}$ which yields the highest information gain over any of the $K$ models. However, such queries are not necessarily helpful to improve the quality of the decision-making. Indeed, they may have little to no impact on the posterior predictive distributions at $\tx$. Moreover, they may improve predictions only for a decision that has very little probability of being the optimal one. Such a phenomenon is displayed on the top-right panel of Figure~\ref{fig-1}.

Thus, we present in the next section a novel active learning strategy, which takes the decision-making problem into account by considering the posterior distribution of the optimal decision for $\tx$, and which therefore overcomes the aforementioned shortcomings.

\section{Targeted active learning criterion} \label{sec-criterion}

\subsection{Posterior uncertainty on the optimal decision} 

The optimal decision is the one with the highest expected utility. If we knew the value of $f_{k,\tx}$ exactly for all $k$, then the optimal decision would be known with $100\%$ certainty. However, since we work with a finite sample size, we cannot have access to the value of $f_{k,\tx}$, but rather characterize posterior distributions $p(f_{k,\tx}|\mathcal{D}_k)$, which in turn leads to the Bayes-optimal recommendation of Eq.~\eqref{eq-dbayes}. As it turns out, by fully taking advantage of this Bayesian framework, we can go beyond that recommendation and come up with the posterior uncertainty that decision $k$ is the optimal decision.

Let us define the random variable $\bar{Y}_k$ as the conditional expectation of $Y_k$ given the random variable $f_{k,\tx} \sim p(f_{k,\tx}|\mathcal{D}_k)$. As it turns out, we have
\begin{equation}
    \bar{Y}_k = f_{k,\tx},
\end{equation}
since we assume the additive noise to be zero-mean. The randomness of $f_{k,\tx}$ (and thus of $\bar{Y}_k$) only comes from lack of information. Such uncertainty is said to be \textit{epistemic}, and adding more points to the dataset will reduce this uncertainty \citep{hullermeier2021aleatoric}.

We can now compute the posterior probability that decision $k$ is the optimal decision, which we denote by $\pi_k$. We have
\begin{align}
    \pi_k & = \mathbb{P} \left( f_{k,\tx} = \max_{k'} f_{k',\tx} \right), \\
    & = \mathbb{P} \left( \bigcap_{k' \neq k} \{ f_{k,\tx} > f_{k',\tx} \} \right). \label{eq-pik}
\end{align}
Note that the events inside Eq.~\eqref{eq-pik} are not independent, and as such this cannot be broken down into a product of probabilities. We denote by $\Db(\tx)$ the discrete random variable which contains the posterior uncertainty on the optimal decision for $\tx$, i.e., whose probability mass function is given by the $(\pi_k)$. By definition we have
$\mathbb{P}(\Db = k) = \pi_k$, and the $\pi_k$ sum to 1, so this defines a valid random variable. An illustrative example is given in the left and center panels of Figure~\ref{fig-1}.

\subsection{Decision-targeted active learning criterion}

\begin{figure}[t]
    \centering
    \includegraphics[width=0.5\columnwidth]{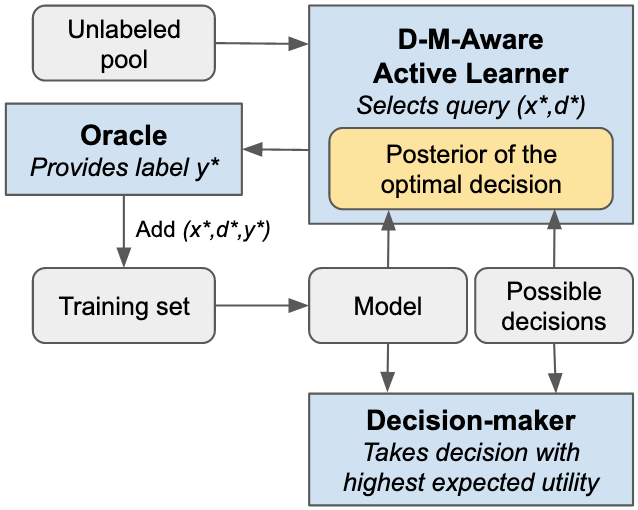}
    \caption{Decision-making-aware active learning. Agents are in blue boxes. The active learner is aware of the the down-the-line decision-making problem, and selects targeted queries for the problem, by taking into account the posterior distribution of the optimal decision. Once the learning phase is over, we here assume that the end-user takes the action which yields the highest expected utility.}
    \label{fig-2}
\end{figure}

Now that we have characterized the posterior distribution of interest (the distribution of the variable we called $\Db(\tx)$), we propose to sequentially select the data point from $\mathcal{U}$ which maximizes the expected information gain about this posterior distribution. We write
\begin{align}
    & (\mathbf{x}^{\star}, d^{\star}) = \argmax_{(\mathbf{x}_j, d_j) \in \mathcal{U}} \bigg( \Hh[p(\Db(\tx)|\mathcal{D})] \label{eq-crit-1} - \mathbb{E}_{p(y_{d_j}|\mathbf{x}_j, \mathcal{D}_{d_j})} \left[ \Hh [ p(\Db(\tx) | \mathcal{D} \cup \{ (\mathbf{x}_j, d_{j}, y_{d_j} ) \} ) ] \right] \bigg),
\end{align}
which means that these queries aim at reducing the uncertainty on the optimal decision of $\tx$. The criterion of Eq.~\eqref{eq-crit-1} may be rewritten in a simpler form as
\begin{align}
    & (\mathbf{x}^{\star}, d^{\star}) = \argmin_{(\mathbf{x}_j, d_j) \in \mathcal{U}} \label{eq-crit-2} \mathbb{E}_{p(y_{d_j}|\mathbf{x}_j, \mathcal{D}_{d_j})} \left[ \Hh [ p(\Db(\tx) | \mathcal{D} \cup \{ (\mathbf{x}_j, d_{j}, y_{d_j} ) \} ) ] \right].
\end{align}
When we are interested in making decisions for a population of test points, Eq.~\eqref{eq-crit-2} simply becomes
\begin{align}
    & (\mathbf{x}^{\star}, d^{\star}) = \argmin_{(\mathbf{x}_j, d_j) \in \mathcal{U}} \label{eq-crit-mult} \mathbb{E}_{p(y_{d_j}|\mathbf{x}_j, \mathcal{D}_{d_j})} \left[ \sum_i \Hh [ p(\Db(\tx_i) | \mathcal{D} \cup \{ (\mathbf{x}_j, d_{j}, y_{d_j} ) \} ) ] \right].
\end{align}
The full decision-making-aware active learning process is illustrated on Figure~\ref{fig-2}.

\subsection{Practical implementation}

Computing the criterion Eq.~\eqref{eq-crit-2} requires to solve two computational challenges:
\begin{enumerate}
    \item The expectation w.r.t. $p(y_{d_j}|\mathbf{x}_j, \mathcal{D}_{d_j})$ is intractable and needs to be approximated;
    \item The probabilities $\pi_d$ are not known in closed form either. They need to be estimated in order to compute the entropy of $\Db(\tx)$.
\end{enumerate}

To approximate the expectation, we resort to Monte Carlo approximation. This means that given $N_s$ samples $y_{d_j}^{(l)}$ drawn from $p(y_{d_j}|\mathbf{x}_j, \mathcal{D}_{d_j})$, we have
\begin{align}
    & \mathbb{E}_{p(y_{d_j}|\mathbf{x}_j, \mathcal{D}_{d_j})} \left[ \Hh [ p(\Db(\tx) | \mathcal{D} \cup \{ (\mathbf{x}_j, d_j, y_{d_j} ) \} ) ] \right] \notag \\
    & \simeq \frac{1}{N_{s}} \sum_{l=1}^{N_{s}} \Hh [ p(\Db(\tx) | \mathcal{D} \cup \{ (\mathbf{x}_j, d_j, y_{d_j}^{(l)} ) \} ) ].
\end{align}
Note that when $p(y_{d_j}|\mathbf{x}_j, \mathcal{D}_{d_j})$ is Gaussian, as is the case in GP regression, we may use a Gauss-Hermite approximation scheme (see Appendix~\ref{app-D}).

Next, to compute the entropy $\Hh [ p( \Db(\tx) | \mathcal{D} ) ]$, we need to know the posterior probabilities $\pi_k$ (given by Eq.~\eqref{eq-pik}). Unfortunately, closed-form solutions do not exist in general. We take sets of posterior draws from $p(f_{k, \tx}|\mathcal{D}_k)$ for all $k$ to generate posterior samples of $\Db(\tx)$, which are then used to estimate the entropy. For simplicity, we estimate the entropy of the multinomial distribution by using empirical estimates of the $\pi_k$ from the posterior samples of $\Db(\tx)$.

Pseudo-code of the algorithm computing the proposed targeted AL criterion is given in Algorithm~\ref{alg-1}. All the computational burden resides in the model retraining step (line 6), which has to be carried out $N_s \times \text{card}(\mathcal{U})$ times to solve Eq.~\eqref{eq-crit-2}. The computational complexity is high, but many operations are trivially parallelizable, for example over all elements of $\mathcal{U}$, or even over all Monte Carlo samples. Moreover, pre-selection strategies may be implemented to avoid computing the criterion for all elements of $\mathcal{U}$, or the selection problem itself could be cast as a Bayesian optimization problem.

\begin{algorithm}[t]
    $C = 0$; \# \textit{Current estimate of Eq.~\eqref{eq-crit-2}} \\
    \# \textit{Monte Carlo approximation} \\
    \For{$l = 1, \dotsc, N_{s}$ }{
        $y_{d_j}^{(l)} \sim p(y_{d_j}|\mathbf{x}_j, \mathcal{D}_{d_j})$; \\
        Add $(\x_j, y_{d_j}^{(l)})$ to the training set $\mathcal{D}_{d_j}$; \\
        Retrain GP associated with decision $d_j$; \\
        \# \textit{Estimate entropy of $\Db(\tx)$ with augmented dataset} \\
        \For{$k = 1, \dotsc, K$}{
            Get samples from $p(f_{k,\tx}|\mathcal{D}_{k})$; \\
        }
        Compute estimates of the $\pi_k$; \\
        Compute entropy $\Hh$ from the $\pi_k$; \\
        $C = C + N_{s}^{-1} \Hh$; \\
        Remove $(\x_j, y_{k,j}^{(l)})$ from the training set $\mathcal{D}_{d_j};$ \\
    }
\caption{Estimating the criterion Eq.~\eqref{eq-crit-2} for $(\x_j, d_j) \in \mathcal{U}$}
\label{alg-1}
\end{algorithm}

Lastly, we emphasize that working with a test population, i.e., dealing with Eq.~\eqref{eq-crit-mult} instead of Eq.~\eqref{eq-crit-2}, brings negligible additional computational complexity. Indeed, the only difference is that we would have to estimate several entropy values instead of one, which has a negligible cost compared to retraining the model (line 6), as previously stated.

\section{Related work} \label{sec-related}

\paragraph{Decision-making-aware strategies in machine learning.}

We begin by discussing such strategies in a \textit{passive} learning context. \citet{lacoste2011approximate} introduced the so-called loss-calibrated inference framework. The decision-making problem is there characterized by a loss (i.e., negative utility), which is taken into account to alter the learning objective of variational inference. This work has been extended, e.g.,  to Bayesian neural networks \citep{cobb2018loss} and to continuous decisions \citep{kusmierczyk2019variational}. Another line of work, which tackles the computation of expected functions (w.r.t. a  posterior distribution), is discussed by \citet{rainforth2020target}. The authors argued that when these functions are known in advance, it is beneficial to take them into account and subsequently proposed a framework coined TABI (target-aware Bayesian inference), which enables efficient estimation of such quantities.

Surprisingly enough, the literature is quite sparse when it comes to similar strategies in for active learning. \cite{saar2007decision} proposed two heuristics to help choosing which customers to target in marketing campaigns. More recently, \citet{sundin2019active} proposed a novel active learning criterion based on the Type-S error to improve binary decisions. Several recent works tackled goal-oriented active learning, but none of those consider the decision-making step that comes after the learning process. For instance, \citet{kandasamy2019myopic} introduced a reward function and a method based posterior sampling, and \citet{xu2019understanding} introduced a utility function and the use of so-called influence functions, but the words ``reward'' or ``utility'' there refer to different metrics of model evaluation. Finally, \citet{zhao2021uncertainty} proposed an uncertainty-aware AL criterion for classification with 0-1 loss, which focuses only on the reduction of the uncertainty that pertains to the classification error.

\paragraph{Bayesian optimization and active learning.}

Bayesian optimization (BO) refers to a class of algorithms for global optimization of black-box functions, where a probabilistic surrogate model such as a Gaussian process is placed on the objective function \citep{Jones1998,brochu2010tutorial}. BO algorithms sequentially select points where the objective function is evaluated, based on some acquisition function which typically balances exploration and exploitation. As such, BO is closely related to AL, see for example \citet{ling2016} for a unifying framework of some standard AL and BO algorithms. Conceptually, BO can be seen as a goal-oriented AL strategy, but for the specific decision-making problem of finding the global optimizer of some black-box function. Only in this setting (and assuming that our framework could be extended to continuous decision variables) would the proposed method amount to BO. In general, there is no obvious formulation of BO which solves our AL problem. If the black-box function was the utility, BO would then find the pair $(\x,d)$ with maximum utility, e.g., a patient with a treatment that has maximum utility in personalized medicine.

\paragraph{Best arm identification in multi-armed bandits.}

The decision-making problem we consider can equivalently be presented as the problem of identifying which of the $K$ arms, described by the distributions of the utilities of each decision at $\tx$, is the best (i.e., yields the highest expected utility, or reward). This is known in the multi-armed bandits literature as the ``best arm identification'' problem, or ``pure exploration'' problem, which has been studied both from frequentist and Bayesian perspectives \citep{audibert2010best,kaufmann2016complexity,russo2016simple}. The objective of such problems differs from the traditional setting of multi-armed bandits, which is to maximize the cumulative sum of rewards.

However, the setting of best arm identification problems differs from ours in the possible arms that can be sampled. In contrast to these problems, we cannot sample from the different arms at $\tx$. We can only sample once from a specific set arms defined by the pairs $(\x,d) \in \mathcal{U}$. This prevents us from using strategies from the multi-armed bandits literature. Instead, by adding new points to the regression models, we aim at better characterizing the distributions of the expected utilities at $\tx$.

\section{Experiments} \label{sec-exp}

\subsection{Use-cases and datasets}

\paragraph{Fully synthetic data.}

We proceed to generate a dataset of 400 points of dimension 5. The covariates are drawn from the standardized Gaussian distribution. We generate four different outcomes as independent realizations of GPs with squared exponential kernels whose variance and lengthscales are different. These outcomes are then corrupted by Gaussian white noise. Finally, the decision variable associated to each point is drawn randomly, but not uniformly, to mimic imbalance in treatment assignment. 

\paragraph{Treatment recommendation.}

The first use-case focuses on the topical personalized medicine research question of using electronic health records (EHR) to augment data from randomized controlled trials (RCT). In this setting, the training set contains individuals $\x$, and the outcome $y$ of the treatment $d$ that they received. In addition, we assume a record of patients and treatments, for which the outcomes can be acquired. An example case is EHR that contain information about prescription of a treatment without follow-up, in which case a new appointment or call needs to be scheduled with the patient in order to acquire the outcome. The objective is to improve the decision of which treatment to give to a new patient $\tilde{\x}$, as in~\citet{sundin2019active}.

Experiments are run on the IHDP dataset\footnote{Available online as part of the supplementary material of \cite{hill2011bayesian}.} \citep{hill2011bayesian}, a semi-synthetic dataset which consists of 747 patients with 25 covariates. The patient covariates come from a real randomized medical study from the 80s, however the outcomes have been artificially generated, implying that all potential outcomes are available. We combine the responses A1, B1 and C1 to obtain a 3-decision problem.

\paragraph{Knee osteoarthritis diagnosis.} The second use-case focuses on symptomatic patients who have a suspicion of knee osteoarthritis (OA) progression in the medial compartment of the right knee. OA is a degenerative disorder of the joints, which reveals itself through symptomatic and structural changes. To date, this disease has no cure, but if detected early, its progression could be slowed down via behavioral interventions~\citep{katz2021diagnosis}. We thus consider the problem of optimizing the diagnostic path for a new patient $\tx$. More precisely, the decision-making problem is to decide when to perform the next follow-up: at 12, 24, 36 or 48 months, or after 48 months. We assume that the doctor is able to query for additional data about previous patients, but that requires a laborious authorization process due to privacy concerns.

We construct a dataset from the Osteoarthritis Initiative (OAI) database\footnote{\url{https://nda.nih.gov/oai/}}, which is a multi-center 10-year observational longitudinal study of 4796 subjects (consent obtained from all the subjects; data are de-identified). After pre-processing, we obtain our final dataset with 8 covariates (clinical data and an initial imaging-based assessment) from 606 patients. The outcome is the joint space width loss over 0.7mm by the time of the follow-up~\citep{neumann2009location,eckstein2015cartilage}. A detailed description of the dataset is given in Appendix~\ref{app-F}.

\subsection{Model of the outcomes}

All experiments are run with GP regression \citep{rasmussen2006gp}, i.e., we assume a zero-mean GP prior for the function $f_k$, with kernel $\kappa_k$:
\begin{equation}
    f_k \sim \text{GP}(\mathbf{0}, \kappa_k (\x, \x')).
\end{equation}
Note that in this case, posterior distributions $p(f_{k,\tx}|\mathcal{D}_k)$ turn out to be Gaussian (standard results are recalled in the Appendix~\ref{app-B}). We use for all models the squared exponential kernel with automatic relevance determination (ARD-SE). GP hyperparameters (variance, lengthscales), as well as the noise variance are estimated with maximum marginal likelihood. Python implementation is carried out with the framework \texttt{GPy}\footnote{\url{https://sheffieldml.github.io/GPy/}} (open-source, under BSD licence).

\subsection{Protocol and evaluation metrics}

Our experimental protocol is as follows: each considered dataset is randomly split into a training set $\mathcal{D}$, query set $\mathcal{U}$, and a single testing point $\tx$. We then proceed to sequentially acquire $N_{\text{acq}}$ points using the proposed Algorithm~\ref{alg-1} and the active learning baselines presented in the next subsection. All experiments are run with $N_{\text{acq}} = 10$.

We track the evolution of two metrics, computed both before the active learning phase and after each acquisition, over $M$ different splits of the original dataset. More precisely, given a split $m$ we track whether the correct decision is returned, with binary accuracy score
    \begin{equation}
        A_m = \mathbb{I}(d^{m}_{\text{BAYES}}, d^{m}_{\star}),
    \end{equation}
where $d^{m}_{\text{BAYES}}$ is the Bayes-optimal decision for $\tx_m$ returned by the model (i.e., according to Eq.~\eqref{eq-dbayes}), and $d^{m}_{\star}$ is the ground truth best decision for $\tx_m$. We have $\mathbb{I}(d_m,d_m^*)=1$ if and only if $d_m=d_m^*$ (and zero otherwise). Our second metric is the entropy of the posterior of the optimal decision of the testing point $\tx_m$
    \begin{equation}
        H_m = \Hh[p(\Db(\tx_m)|\mathcal{D}_m)].
    \end{equation}
All experiments are run with $M=200$ replications.

\subsection{Baseline active learning methods}

The proposed method (acronym \texttt{D-EIG}) is compared with several active learning methods:

\begin{itemize}
    \item Random sampling (\texttt{RS}) -- Chooses $(\x_j, d_j)$ uniformly at random from $\mathcal{U}$;
    \item Uncertainty sampling (\texttt{US}) -- Chooses the $(\x_j, d_j)$ whose posterior predictive distribution $p(y_{d_j}|\x_j, \mathcal{D}_{d_j})$ has the greatest variance;
    \item Expected information gain (\texttt{EIG}) -- Presented in Section~\ref{sec-bal}. Chooses the $(\x_j, d_j)$ which yields the greatest expected information gain on its associated GP.
    \item Decision uncertainty sampling (\texttt{D-US}) -- A baseline that we introduce. Chooses the $(\x_j, d_j)$ whose optimal decision (i.e., associated to $\x_j$) is the most uncertain, evaluated with the entropy of $\Db(\x_j)$;
    \item Targeted information (\texttt{T-EIG}) -- A targeted AL baseline introduced by \citet{sundin2018improving}. Chooses the $(\x_j, d_j)$ which yields the greatest expected information gain on $p(\ty_{d_j}|\tx, \mathcal{D}_{d_j})$. This criterion is connected to the classical expected error reduction criterion, see details in Appendix~\ref{app-G}.
\end{itemize}

\subsection{Results}

\begin{figure}[t]
    \centering
    \begin{tabular}{ccc}
    \includegraphics[width = 0.3\textwidth]{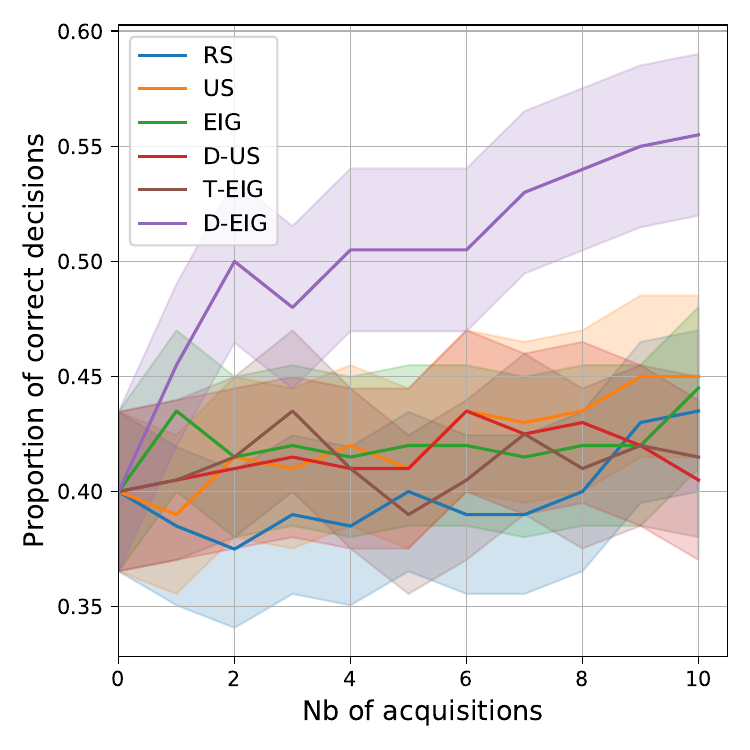}
    &
    \includegraphics[width = 0.3\textwidth]{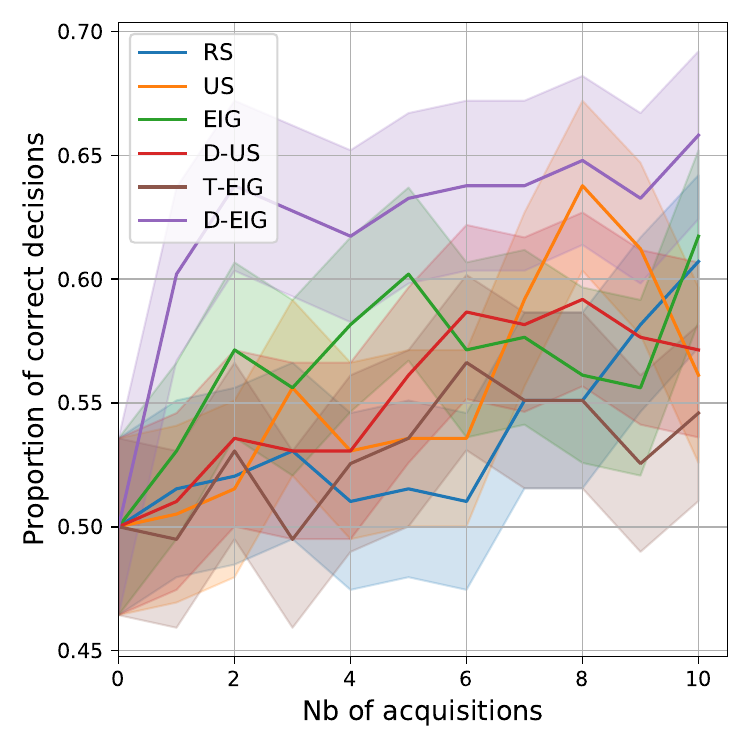}
    &
    \includegraphics[width = 0.3\textwidth]{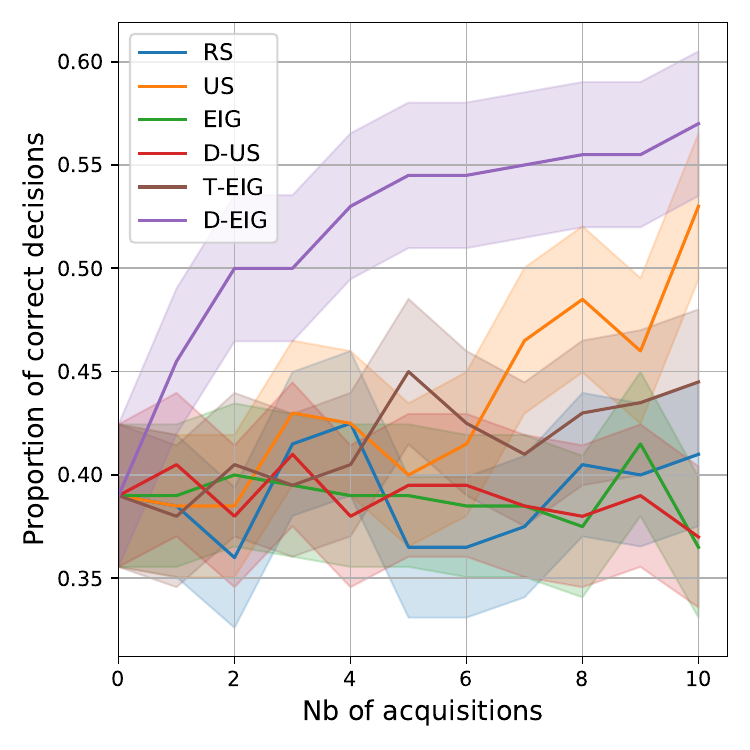} \\
    (a) & (b) & (c)
    \end{tabular}
    \caption{Mean and standard error of the mean of the accuracy score $A_m$ over $M = 200$ replications of the experiment w.r.t. the number of AL acquisitions. The proposed targeted active learning criterion \texttt{D-EIG} outperforms all considered AL methods in improving the accuracy of the decision-making. From left to right: (a) Synthetic data. (b) IHDP dataset. (c) OAI dataset. }
    \label{fig-3}
\end{figure}

\begin{figure}[t]
    \centering
    \begin{tabular}{ccc}
    \includegraphics[width = 0.3\textwidth]{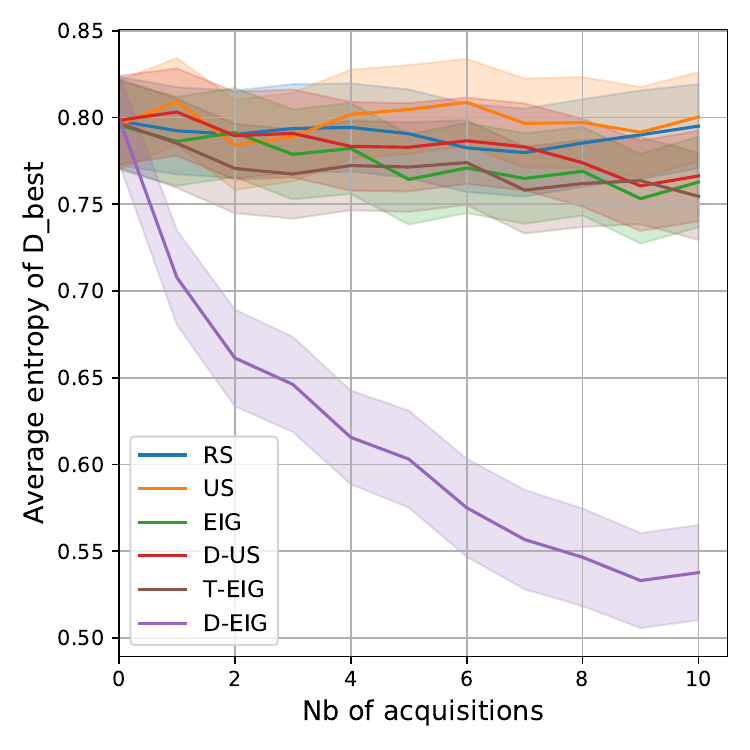}
    &
    \includegraphics[width = 0.3\textwidth]{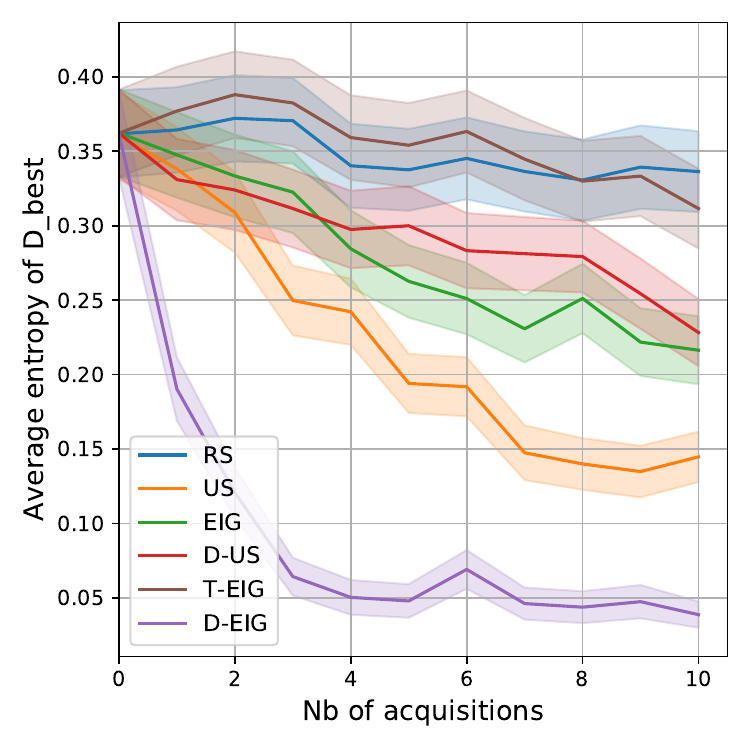}
    &
    \includegraphics[width = 0.3\textwidth]{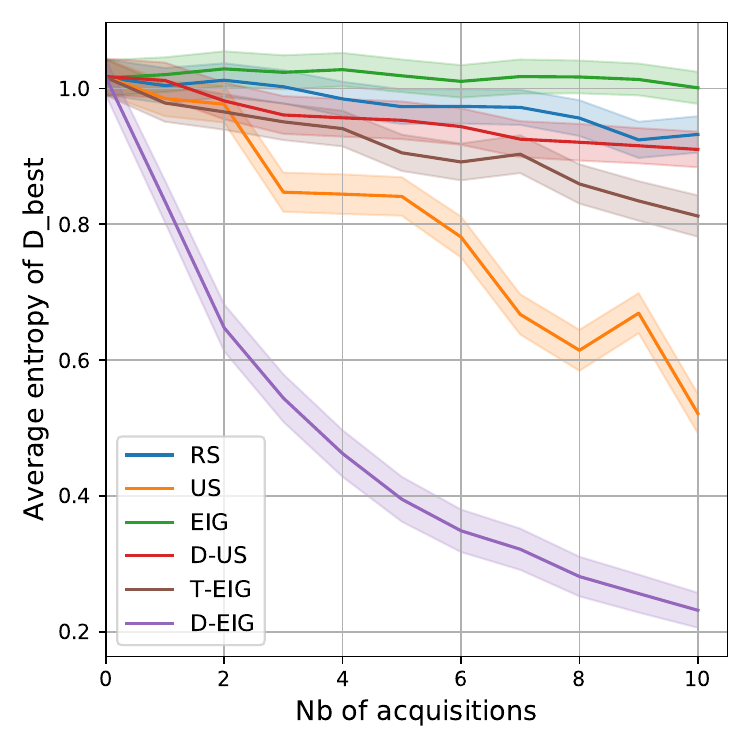} \\
    (a) & (b) & (c)
    \end{tabular}
    \caption{Mean and standard error of the mean of the entropy score $H_m$ (entropy of the posterior of the optimal decision) over $M = 200$ replications of the experiment w.r.t. the number of AL acquisitions. The proposed targeted active learning criterion \texttt{D-EIG} reduces the uncertainty on the optimal the fastest among all considered AL methods. From left to right: (a) Synthetic data. (b) IHDP dataset. (c) OAI dataset.}
    \label{fig-4}
\end{figure}

Experiments are run with a starting training set of size 100 for the synthetic dataset and the OAI dataset, and of size 50 for the IHDP dataset. All experiments were run on a high-performance computing cluster. Figure~\ref{fig-3} displays the evolution of the average binary accuracy score $A_m$ over all replications (i.e., the evolution of the proportion of correct decisions). Figure~\ref{fig-4} displays the evolution of the average entropy of the posterior of the optimal decision (the $H_m$ score). Shaded areas correspond to the standard error of the mean.

\paragraph{Performance of the proposed method.}

As can be seen in the plots, for all considered datasets, the proposed method gives the best results both in terms of improving the decision-making accuracy and reducing the uncertainty on the optimal decision. This is particularly striking in the OAI dataset, where the problem is the hardest (real data and five possible decisions): all alternatives barely improve the decision-making, whereas the proposed method greatly improved it.

\paragraph{Performance of non-targeted baselines.}

As expected, the considered baselines that are not targeted to $\tx$ (namely \texttt{RS}, \texttt{US}, \texttt{EIG}, and \texttt{D-US}) do not yield good performance, with the notable exception of \texttt{US} which has the second-best performance in entropy reduction on the IHDP and OAI datasets.

\paragraph{Performance of the targeted baseline.}

Lastly, the baseline targeted to $\tx$ that we consider, \texttt{T-EIG}, has overall poor performance. This demonstrates the value of taking into account the posterior uncertainty on the optimal decision.

\clearpage

\section{Discussion} \label{sec-ccl}

In this paper, we tackled the problem of decision-making-aware active learning, that is, sample-efficient performance improvement in a down-the-line decision-making problem. To this end, we have proposed a novel criterion, which directly aims at reducing the uncertainty on the posterior distribution of the optimal decision. With experiments we demonstrated the advantages of the proposed technique compared to existing active learning baseline methods in personalized medicine settings. 

The main limitation of the proposed method is its computational complexity, as the current implementation involves many model retraining steps. Computational complexity is tolerable in applications where both the utility of correct decisions and cost of acquiring novel data points are high, such as in personalized medicine. Nevertheless, future work is needed to design lower-complexity and still accurate approximations of the proposed criterion. Extending the proposed criterion to batch selection, in contrast with the current sequential selection method, will also help. The second limitation of our method is that we considered decisions to be available to the algorithm, and in many real-life situations this may not be the case, for instance due to privacy concerns. However, our criterion can be straightforwardly extended to tackle this limitation, and we also see this as a direction for the future work. 

To conclude, we anticipate that our method will have a significant impact in interactive AI with healthcare applications. Specifically, we have shown that the proposed technique can be applied in personalized diagnosis and treatment applications. Both of these clinical problems require accurate and reliable decision-making tools, which are, however, costly to build. Our method is sample-efficient, and has decision-making capabilities by design.

\clearpage

\begin{appendices}

\section{Entropy of a probability distribution} \label{app-A}

The entropy of a probability distribution is a non-negative measure of uncertainty. The higher the value, the more uncertain the outcome is.

\subsection*{Discrete case}

Let $p(z)$ be the p.m.f. of the random variable $Z$, with possible $K$ outcomes which occur with probabilities $\pi_1, \dotsc, \pi_K$. Then
\begin{equation}
    \Hh[p(z)] = - \sum_{k=1}^{K} \pi_k \log \pi_k.
\end{equation}

\subsection*{Continuous case (differential entropy)}

Let $p(z)$ be the p.d.f. of the random variable $Z$ with support $\mathcal{Z}$. Then
\begin{equation}
    \Hh[p(z)] = - \int_{\mathcal{Z}} p(z) \log p(z) \dd z.
\end{equation}

\subsection*{Entropy of a Gaussian random variable}

The entropy of a Gaussian random variable with mean $\mu$ and variance $\sigma^2$ is equal to $\frac{1}{2} \log(2\pi e \sigma^2)$.

\section{Gaussian process regression} \label{app-B}

The elements of this section are taken from \citet{rasmussen2006gp}, Chapter 2.

A Gaussian process (GP) is a stochastic process, i.e., a collection of random variables, such that any finite combination of this collection has a Gaussian distribution. A GP is completely specified by its mean function $m$, and covariance function (or kernel) $\kappa$, and we write
\begin{equation}
    f \sim \text{GP}(m(\x), \kappa(\x, \x')).
\end{equation}
It is often assumed that $m(\x) = 0$.

We now consider a GP regression model
\begin{align}
    f & \sim \text{GP}(\mathbf{0}, \kappa(\x, \x')), \\
    y & = f(x) + \epsilon,
\end{align}
where $\epsilon \sim \mathcal{N}(0,\sigma^2)$. That is to say that a GP prior is placed on $f$. In the following, we use the notation $f_{\x})$ to denote $f(\x)$. Given a collection of observations $\mathcal{D} = \{\x_i, y_i \}_{i=1}^{N}$, we wish to characterize the posterior distribution at a test point $\tx$, $p(f_{\tx}|\mathcal{D})$. The definition of a GP implies that
\begin{equation}
    \begin{bmatrix}
    \mathbf{y} \\ f_{\tx}
    \end{bmatrix}
    \sim \mathcal{N} \left(
    \mathbf{0},
    \begin{bmatrix}
    \kappa(\mathbf{X}, \mathbf{X}) + \sigma^2 \mathbf{I} & \kappa(\mathbf{X}, \tx) \\
    \kappa(\tx, \mathbf{X}) & \kappa(\tx, \tx)
    \end{bmatrix}
    \right),
\end{equation}
and as such, by using basic manipulations of the Gaussian distribution, it can be shown that
\begin{equation}
    p(f_{\tx}|\mathcal{D}) = \mathcal{N}(\mu_{\tx}, \sigma^2_{\tx}),
\end{equation}
where
\begin{align}
    \mu_{\tx} & = \kappa(\tx, \mathbf{X}) [\kappa(\mathbf{X}, \mathbf{X}) + \sigma^2 \mathbf{I}]^{-1} \mathbf{y} \\
    \sigma^2_{\tx} & = \kappa(\tx, \tx) - \kappa(\tx, \mathbf{X}) [\kappa(\mathbf{X}, \mathbf{X}) + \sigma^2, \mathbf{I}]^{-1} \kappa(\mathbf{X}, \tx).
\end{align}
Consequently, $p(y|\tx, \mathcal{D})$ is also Gaussian with mean $\mu_{\tx}$ and variance $\sigma^2 + \sigma^2_{\tx}$.

\section{Notes on the expected information gain (EIG)} \label{app-C}

\subsection*{EIG and mutual information}

Let us consider a standard Bayesian regression model, with likelihood $p(y|\x,\theta)$ and prior $p(\theta)$, which leads to the characterization of the posterior distribution $p(\theta|\mathcal{D})$. The information brought by a new observation $(\x, y)$ is the reduction in posterior entropy
\begin{equation}
    \text{IG}(\x,y) = \Hh[p(\theta|\mathcal{D})] - \Hh[p(\theta|\mathcal{D} \cup \{ (\x,y) \})].
\end{equation}
However, in practice, we only have access to $\x$ and are uncertain about the outcome. Therefore we define the expected information gain brought by $\x$ as the expectation of the previous expression w.r.t. the posterior predictive distribution $p(y|\x, \mathcal{D})$:
\begin{equation}
    \text{EIG}(\x) = \Hh[p(\theta|\mathcal{D})] - \mathbb{E}_{p(y|\x, \mathcal{D})} \big[ \Hh[p(\theta|\mathcal{D} \cup \{ (\x, y) \}) ] \big].
\end{equation}

This expression can be rearranged to show that the EIG is equal to the mutual information between $y$ and $\theta$ (given $\x$ and $\mathcal{D}$), defined as
\begin{equation}
    \text{I}(y;\theta|\x,\mathcal{D}) = \iint p(y,\theta|\x,\mathcal{D}) \log \frac{p(y,\theta|\x,\mathcal{D})}{p(y|\x,\mathcal{D}) p(\theta|\x,\mathcal{D})} \dd y \dd \theta.
\end{equation}
The symmetry of the mutual information leads in turn to an alternative formulation of the EIG, namely
\begin{equation}
    \text{EIG}(\x) = \Hh[p(y|\x,\mathcal{D})] - \mathbb{E}_{p(\theta|\mathcal{D})} [ \Hh [ p( y | \x, \theta) ] ].
    \label{eq-supp-eig}
\end{equation}
which now computes entropies in the output space (and not the parameter space). Most notably, this does not involve model retraining. This is the form most often used in practice.

\subsection*{Extension to non-parametric models}

If we now consider a regression model of the form $y = f(\x) + \epsilon$, where a functional prior is placed on $f$, and $\epsilon \sim \mathcal{N}(0, \sigma^2)$, we can adapt the expression of Eq.~\eqref{eq-supp-eig} as
\begin{equation}
    \text{EIG}(x) = \Hh[p(y|\x, \mathcal{D})] - \mathbb{E}_{p(f|\mathcal{D})} \big[ \Hh[p(y|\x, f) ] \big].
    \label{eq-supp-eig-gp}
\end{equation}

\subsection*{Special case for Gaussian process}

When dealing with GP regression, the predictive posterior distribution $p(y|\x,\mathcal{D})$ is Gaussian, with mean $\mu_{\x}$ and variance $\sigma^2 + \sigma^2_{\x}$. Considering that the value of $\sigma^2$ is fixed, or estimated, the expression of Eq.~\eqref{eq-supp-eig-gp} becomes
\begin{equation}
    \text{EIG}({\x}) = \frac{1}{2} \left( \log(\sigma_{\x}^2 + \sigma^2) - \log(\sigma^2) \right).
\end{equation}
As such, the higher $\sigma_{\x}^2$, the higher the EIG.

\section{Gauss-Hermite quadrature} \label{app-D}

We consider computing expectations of the form
\begin{equation}
    \mathbb{E}[f(y)] = \int f(y) p(y) \dd y,
\end{equation}
where $Y$ is a Gaussian random variable with mean $\mu$ and variance $\sigma^2$. The Gauss-Hermite approximation of order $N$ of the previous expression is given by
\begin{equation}
    \frac{1}{\sqrt{\pi}} \sum_{i=1}^{N} \omega_i f(\sqrt{2}\sigma x_i + \mu),
\end{equation}
where the $x_i$ are the roots of the Hermite polynomial of order $N$ (denoted by $H_n$), and the weights $\omega_i$ are given by
\begin{equation}
    \frac{2^{n-1} n! \sqrt{\pi}}{n^2 (H_{n-1}(x_i))^2}.
\end{equation}

\section{Extension to a test population} \label{app-E}

We now consider instead a collection of previously unseen testing points $\tx_i~(i \in \{1, \dotsc, N_{\text{test}} \})$, for which the end-user has to make decisions, i.e. choose for each of these points one action among a set of $K$ available ones. The optimal decision for $\tx_i$ is to be understood, as before, as the one which yields the highest expected utility at $\tx_i$. Similarly to the developments of Section 3.1, we can define $\Db(\tx_i)$ the discrete r.v. which contains the (epistemic) posterior uncertainty on the optimal decision for each $\x_i$. 

Then, to extend the proposed AL criterion defined for a single $\tx$, we simply consider the expected information gain brought by a new point $(\x, d)$ over the \emph{sum of entropies} of the posteriors of the optimal decision. Removing constant terms, we end up with the following criterion
\begin{align}
    (\mathbf{x}^{\star}, d^{\star}) = \argmin_{(\mathbf{x}_j, d_j) \in \mathcal{U}} \mathbb{E}_{p(y_{d_j}|\mathbf{x}_j, \mathcal{D}_{d_j})} \left[ \sum_i \Hh [ p(\Db(\tx_i) | \mathcal{D} \cup \{ (\mathbf{x}_j, d_{j}, y_{d_j} ) \} ) ] \right].
\end{align}

\section{Knee osteoarthritis follow-up data details} \label{app-F}

We considered all the data from the Osteoarthritis Initiative Dataset (OAI; \url{https://nda.nih.gov/oai/}) with total WOMAC score  over 9 (symptomatic subjects). Subsequently, we selected those subjects, which have early, doubtful, or early radiographic Osteoarthritis at the baseline according to the Kellgren Lawrence grading scoring system.

In our experiments, we used a commonly accepted measure -- joint space width (JSW) loss over $0.7mm$ as an indicator of progression. The JSW was measured from knee X-rays at a fixed location ($x=0.250$), thus focusing on OA only in the medial compartment of the knee.

The following is the list of variables, which we selected from the OAI dataset (per knee):

\begin{itemize}
    \item Age;
    \item Sex;
    \item Body-mass-index (BMI);
    \item Total WOMAC score;
    \item Indication of varus, valgus, or neither;
    \item Indication of past injury;
    \item Indication of past surgery;
    \item Kellgren-Lawrence grade;
    \item JSW at fixed location ($x=0.250$).
\end{itemize}

\section{T-EIG and expected error reduction} \label{app-G}

Let us consider that the error measure is the log-loss. Then, in a Bayesian AL framework, the expected error reduction query writes
\begin{align}
    \x_{\text{EER}} = \argmin_{\x \in \mathcal{U}} \mathbb{E}_{p(y|\x,\mathcal{D})} \left[ \sum_{x_j \in \mathcal{X}_t} \Hh [p(y_j | x_j, \mathcal{D} \cup \{ ( \x, y ) \} ) ] \right],
\end{align}
where $\mathcal{X}_t$ is a test population. In the setting considered in the paper, we have $\mathcal{X}_t = \{ \tx \} $, which reduces to
\begin{equation}
    \x_{\text{EER}} = \argmin_{\x \in \mathcal{U}} \mathbb{E}_{p(y|\x,\mathcal{D})} \left[ \Hh [p(\ty | \tx, \mathcal{D} \cup \{ ( \x, y ) \} ) ] \right],
\end{equation}
which is exactly the targeted expected information gain criterion (\texttt{T-EIG}) introduced in \citet{sundin2018improving}.

\end{appendices}

\clearpage

\bibliography{biblio.bib}
\bibliographystyle{apalike}

\end{document}